\documentclass{article}

\usepackage{PRIMEarxiv}

\usepackage[utf8]{inputenc} 
\usepackage[T1]{fontenc}    
\usepackage{hyperref}       
\usepackage{url}            
\usepackage{booktabs}       
\usepackage{amsfonts}       
\usepackage{nicefrac}       
\usepackage{microtype}      
\usepackage{lipsum}
\usepackage{fancyhdr}       
\usepackage{graphicx}       
\graphicspath{{media/}}     
\usepackage{float}
\usepackage{amsmath}
\usepackage{amssymb}
\usepackage{multirow}
\usepackage{multicol}
\usepackage{makecell}
\usepackage{boldline}
\usepackage{subfig}
\usepackage{authblk}

\newenvironment{widefigure}
  {\begin{figure}}
  {\end{figure}}

\newenvironment{widetable}
  {\begin{table}}
  {\end{table}}

\usepackage{xcolor, soul}
\sethlcolor{yellow}

\label{sec:info}

\title{Topology-Agnostic Graph U-Nets for Scalar Field Prediction on Unstructured Meshes}

\author[1]{Kevin Ferguson}
\author[1]{Yu-hsuan Chen}
\author[1]{Yiming Chen}
\author[2]{Andrew Gillman}
\author[2]{James Hardin}
\author[1]{Levent Burak Kara}

\affil[1]{Carnegie Mellon University\\
  5000 Forbes Avenue\\
  Pittsburgh, PA 15213\thanks{Address all correspondence to lkara@cmu.edu}}
\affil[2]{Air Force Research Laboratory\\
  2977 Hobson Way\\
  Wright-Patterson Air Force Base, OH 45433}

\begin{document}
\maketitle

\label{sec:abstract}
\begin{abstract}
Machine-learned surrogate models to accelerate lengthy computer simulations are becoming increasingly important as engineers look to streamline the product design cycle. In many cases, these approaches offer the ability to predict relevant quantities throughout a geometry, but place constraints on the form of the input data. In a world of diverse data types, a preferred approach would not restrict the input to a particular structure. In this paper, we propose Topology-Agnostic Graph U-Net (TAG U-Net), a graph convolutional network that can be trained to input any mesh or graph structure and output a prediction of a target scalar field at each node. The model constructs coarsened versions of each input graph and performs a set of convolution and pooling operations to predict the node-wise outputs on the original graph. By training on a diverse set of shapes, the model can make strong predictions, even for shapes unlike those seen during training. A 3-D additive manufacturing dataset is presented, containing Laser Powder Bed Fusion simulation results for thousands of parts. The model is demonstrated on this dataset, and it performs well, predicting both 2-D and 3-D scalar fields with a median $R^2 > 0.85$ on test geometries. Code and datasets are available at: \url{https://github.com/kevinferg/graph-field-prediction}.
\end{abstract}


\section{Introduction}
\label{sec:intro}
One of the primary challenges of the engineering design process is minimizing the time between iterations. Often, an engineer must perform a Finite Element Analysis (FEA) of a part and evaluate the results before updating the design and repeating the process \cite{cad-fea-1992, iterative-design-1994, structural-optimization-1997}.

Recent advances in computational technology, including improved hardware \cite{hardware-accelerated-2014}, more efficient methods \cite{inherent-strain-2023, modified-inherent-strain-2024}, and the increasing availability of cloud computing resources \cite{cloud-compute-2012}, begin to alleviate this issue by offering several ways to speed up the simulation process. However, lengthy compute times are still a bottleneck in the design pipeline, especially at early stages, when high-fidelity analyses can be a waste of valuable time and resources \cite{design-optimization-1998, design-process-2005}. To this end, data-driven surrogate models have emerged in the engineering world to reduce time between iterations. Due to their ability to universally approximate high-dimensional mappings, neural networks are pervasive among modern machine-learned surrogate models \cite{stressnet-2019, stressgan-2021, truss-stress-2018, truss-transfer-2021, topopt-surrogate-2021, sla-stress-2019, chen2023automating, chen2024bignet, chen2024virl}.

Image-based data in particular is of particular importance to real-world systems. For such problems, convolutional neural networks (CNNs) provide the ability to characterize local relationships in a way that is physically meaningful. That is, convolution harnesses spatial information throughout an image, so that neural network models can recognize familiar patterns in groups of pixels, allowing them to classify or segment image data according to what is depicted in the images \cite{beyond-euclidean-2017}.

CNNs can classify images, but making nodal predictions, such as for segmentation tasks, is often more challenging; for problems like this, the go-to architecture is a U-Net \cite{unet-2015}, which uses not only convolution, but also pooling and un-pooling. These provide the network the ability to capture long-range patterns in an image without requiring a prohibitively large number of successive convolutions. Inputting an image into a U-Net, the output is a segmentation map, classifying each pixel of the input according to what it represents.

While CNNs like U-Net are powerful, their underlying mechanisms are most naturally suited to images. For many engineering analyses involving design of parts, image data is too restrictive to accurately represent 2-D or 3-D shapes. Images (or $n$-dimensional grid-based data, more generally), cannot represent any shape with arbitrary resolution. For this reason, alternative geometric representations have been used for field prediction, such as point clouds \cite{pointnet-2017, pointnet2-2017, flow-point-cloud-2021} and implicit neural representations \cite{mesh-agnostic-pde-2022, time-dependent-neural-2023}. However, by far the most ubiquitous way to represent arbitrary geometries is with mesh data structures \cite{mesh-based-geometry-2003}. Meshes can be generated with a variety of methods, and they can represent non-uniform detail throughout space, contrary to a grid-based representation. This non-uniformity, however, means that a typical CNN-based U-Net approach will not apply, requiring, for example, a more general \textit{graph} neural network (GNN) to accommodate the unstructured data type.

Our proposed solution is Topology-Agnostic Graph U-Net (TAG U-Net), a novel interpretation of U-Net for graphs, which uses $k$-d tree pooling, a simple yet effective pooling strategy, and EdgeConv \cite{edgeconv-2019} for convolution. Its flexibility to input graph structure makes our method applicable in many contexts for any type of $n$-dimensional geometric data. The model accepts a shape as input, consisting solely of a set of nodal coordinates and edge connectivity information; it outputs the predicted value of the target field at every node.

In this work, we focus on prediction of physical scalar fields in engineering problems. Specifically, we examine a 2-D structural analysis problem, in which the goal is to predict the von Mises stress field. Next we use our method to predict z-displacement in a 3-D additive manufacturing (AM) problem. These test problems are selected for their relevance in identifying issues with proposed part designs. For example, high values of von Mises stress is an indicator of material yield, while z-displacement values above a threshold in metal AM methods, e.g. Laser Powder Bed Fusion (LPBF), may result in recoater blade interference, hence a failed build process \cite{mpbf-modeling-2022}. Being able to predict these fields instantaneously, rather than having to run FEA or simulation, will save significant time for an engineer designing a part. Our 3-D dataset of AM simulation results has been made publicly available for benchmarking in graph network research and for general utility in the AM community.

We trained our proposed model on these datasets, comparing the results with those of a pure GNN model without a U-Net-inspired architecture. We also compared among several model sizes, to determine the performance improvement one can expect by increasing the size of their network.

The main contributions of this work are as follows:
\begin{enumerate}
    \item A 3-D dataset of simulated additive manufacturing results on a large set of shapes
    \item A method for coarsening meshes using $k$-d trees for use in graph networks, and a corresponding pooling operation
    \item A Topology-Agnostic Graph U-Net (TAG U-Net) architecture for use on meshes with varying topologies
    \item A demonstration of a TAG U-Net in two engineering settings: 2D plane-strain stress prediction, and 3D additive manufacturing displacement prediction
\end{enumerate}

\section{Related Work}
\label{sec:related}
GNNs have long been of interest due to how they can make predictions on unstructured data \cite{gnn-methods-2020, gnn-2021}. Specifically, convolutional GNNs mirror the ideas behind image-based CNNs\cite{GCNConv-2017} methods and have been a popular area of research since. GNNs have been used to great success for predicting node values in physical problems, especially in time-evolving systems. For example, Pfaff et al. predict the node location updates in mesh-based physics simulations \cite{meshsimulations-2021}. They have further been employed for time-independent predictions as well, such as fluid dynamics \cite{direct-fluid-2021}, structural analysis \cite{truss-transfer-2021}, and materials engineering \cite{microstructure-2022}.

One pervasive issue among GNNs is that as the number of successive convolutions applied to a graph increases (that is, for increasing depth), performance often falters \cite{oversmoothing-2020}. High depth is desirable, since feature information is allowed to propagate further across a graph. However, this has been known to cause problems with vanishing gradients during training and over-smoothing in predictions. Significant efforts have been made to allow GNNs to be constructed with large depths without a reduction in performance, such as DeepGCN and DeeperGCN by Li et al. \cite{deepgcn-2019, deepergcn-2020}, which demonstrate a set of strategies for training GNNs with high depth. Zhao and Akoglu target the over-smoothing issue in particular by incorporating a specialized normalization function \cite{oversmoothing-2020}.

One other way to target these depth-based GNN issues is to borrow key ideas from image-based CNNs. When learning on grids of data, such as images in 2D or voxel arrays in 3D, the CNN architecture that has become the standard is U-Net \cite{unet-2015}, which combines convolution with image pooling in an encoder-decoder structure, utilizing skip-connections to preserve high-resolution detail across the bottleneck. Such an architecture is effective at performing image-to-image mappings for the purpose of segmentation. It stands to reason, then, that a GNN with similar structure and analogous components, i.e. a Graph U-Net, would perform well at graph-to-graph mapping tasks.

Graph U-Nets like this were originally investigated by Gao and Ji \cite{graphunet-2019}. They explore the use of top-k pooling \cite{top-k-pooling-2018} as an analog of image pooling, and demonstrate strong classification results on citation network datasets. This laid a good foundation for a U-Net like encoder-decoder structure in GNNs.

Some Graph U-Nets on meshes have been proposed for use on single meshes. That is, by defining a pooling scheme that works for one mesh, this can be applied models for any field prediction problem on that mesh. Deshpande et al. \cite{magnet-2024} investigate such a technique for 3-D stress prediction in biological 3-D structures. They demonstrate that this U-Net ideology extends to volumetric meshes. However, a trained model can only make predictions on a single mesh topology. A topology-agnostic approach would allow for predictions to be made on unseen shapes, which is critical for a data-driven model used for design. In the context of surface meshes specifically, MeshCNN, presented  by Hanocka, et al. \cite{meshcnn-2019} provides a solution to several learning tasks in 3-D. While this model does extend to other topologies, it uses pooling and convolution methods specifically designed for triangular surface meshes. The model presented in this work, can operate on mesh structures, and it is demonstrated on two different types of meshes. Another approach relying on GNNs with an encoder-decoder structure for time-independent PDEs is presented by Gladstone et al. \cite{time-independent-2024}; by augmenting the input mesh with additional long-range edges, they achieve good results for stress and distortion predictions, although this is demonstrated only for 2-D input meshes.

Fundamental to our TAG U-Net model is a flexible approach to pooling, one of the major steps in an image U-Net that needs a graph-based analog. The pooling strategy we choose is based on the $k$-d tree of nodes in the mesh. Use of $k$-d trees has appeared in neural network models before. For example, Klokov and Lempitsky demonstrate segmentation on 3-D point clouds using a method based on the $k$-d tree of points \cite{escape-2017}. Furthermore, tree-based representations can be used to accelerate simulations, for instance Li et al., who use coarsening based on the related octree data structure \cite{octree-2019} to accelerate the simulation of AM processes. More details on how $k$-d trees are used in our method are found in Sec.\ref{sec:pooling}.

\section{Methods}
\label{sec:methods}

\subsection{Datasets}
\label{sec:datasets}

We wish to verify that our approach is appropriate in both 2-D and 3-D settings. Therefore, we study two scalar field prediction problems, one in 2-D and one in 3-D. These problems are intended to be examples of contexts in which such a prediction method would be most useful in engineering design and/or optimization. Example shapes from each dataset and their corresponding scalar fields can be found in the Appendix.  


\subsubsection{2-D Dataset: von Mises Stress of a Compressed Part}

\begin{widefigure}[htbp]
    \centering
    \includegraphics[width=0.95\textwidth]{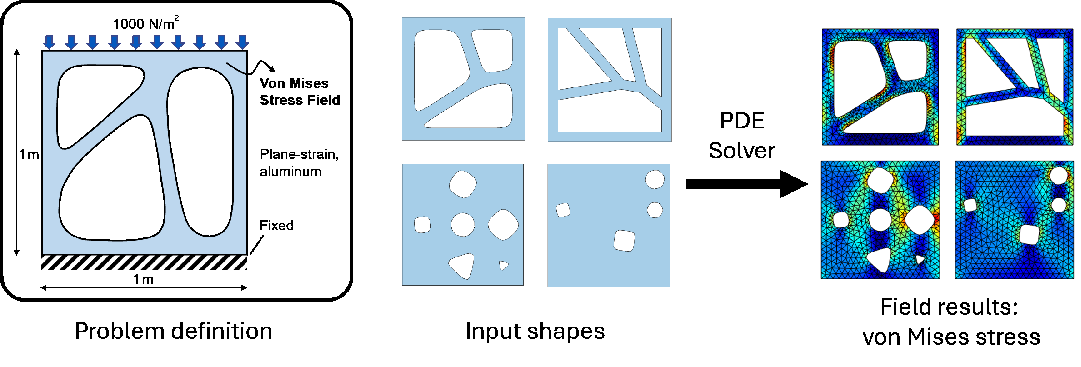}
    \caption{2-D dataset generation process: shapes are inputted into a PDE solver; loads, boundary conditions, and material properties are assigned as in the problem definition shown; von Mises stress fields are outputted as mesh data}
    \label{fig:2d-dataset}
\end{widefigure}

The 2-D dataset, from \cite{sse-2022} and \cite{multires-2024}, consists of square geometries with pores of random shapes and arrangements. Each shape contains the same square outer hull, but the internal pores contain significant variation from shape to shape. The target field to predict for this dataset is von Mises stress, assuming the part is treated as plane-strain, undergoing uniform compression from above, with a fixed bottom boundary and the material properties of aluminum. The dataset generation process is illustrated in Fig.\ref{fig:2d-dataset}. This stress field could, in practice, be used to detect where the part may yield (i.e. where von Mises stress exceeds a threshold). This dataset also contains the signed distance field (SDF) value at each node, which encodes the Euclidean distance from a boundary of the shape \cite{SDF-2003}. During prediction, we augment the nodal input feature set to include this SDF feature in addition to nodal coordinates. Meshes in this dataset contain triangular elements.

This dataset contains 2,000 total shapes. 1600 are used for training, while the last 400 are unseen during training but used for testing the trained model. Several example shapes from this dataset and their von Mises stress results are included in the Appendix for reference. Note that the shared square boundary in this dataset gives it less shape diversity than the 3-D dataset, which will be described shortly. However, due to the complexity of its von Mises stress fields and the ubiquity of stress analyses in engineering, we find this a suitable dataset for testing our methods.

\subsubsection{3-D Dataset: Simulated Additive Manufacturing Displacement Fields}

We introduce a new 3-D dataset to test our method in a real design domain. To obtain a large corpus of shapes, we look to the Fusion 360 Gallery Segmentation Dataset \cite{fusion360-2021}. We target this dataset due to its availability and due to the wide variety of shapes it contains. We simulate the assembly of each part via Laser Powder Bed Fusion (LPBF) using Autodesk Netfabb Simulation Utility \cite{netfabb-2024}. This performs multiple finite element analyses on the part, generating both a thermal and a mechanical history during the build process. We scale the Fusion 360 part by $10\times$ and translate the part such that it rests on the build surface at $z=0~\text{mm}$; the parts are not otherwise rotated or transformed. The simulation is then carried out assuming the machine is a Renishaw AM250 and the material is Inconel 625, 40 \textmu m thickness. Other parameters are left as their default nominal values and no support structures are added. The simulation software re-meshes the geometry to contain 8-node axis-aligned hexahedral elements before simulating the build process. The dataset generation process is outlined in Fig.\ref{fig:3d-dataset}.

\begin{widefigure}[htbp]
    \centering
    \includegraphics[width=0.95\textwidth]{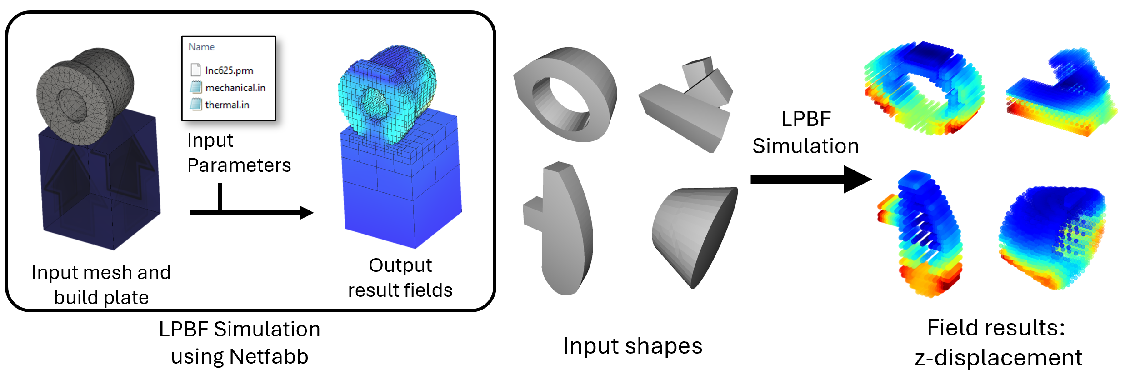}
    \caption{3-D dataset generation process: shapes are inputted into Netfabb; material properties, and process parameters are assigned via input files; z-displacement fields are outputted as mesh data}
    \label{fig:3d-dataset}
\end{widefigure}

The mechanical simulation results contain several field quantities, tracked throughout the part at several time steps during the build. These nodal quantities are: displacement (3 components), elastic strain (6 components), Cauchy stress (6 components), von Mises stress (scalar), and 3 principal stresses (1 component each) with their directions (3 components each). We make available the full dataset including all nodal field results along with a small suite of functions for manipulating and visualizing the data. We also provide functions for loading the data as a PyTorch Geometric \texttt{Data()} graph \cite{pytorch-2017}, for use in graph neural network training. 

To test our model, we focus on one field prediction problem that is highly relevant to our dataset and the AM community in general: prediction of residual nodal vertical displacement. A way to quickly predict displacement before a build is highly sought-after for rapid design troubleshooting purposes. For one, a highly distorted part will not be usable after manufacturing. Additionally, knowing nodal displacement ahead of time can also help predict whether a build may fail. In metal additive manufacturing processes like LPBF, after the laser's path solidifying a layer of material on the part, a recoater blade helps to evenly distribute powder across the build area \cite{recoater-effect-2024}. Overhanging features can lead to displacement in the (vertical) z-direction during the build process, which can interfere with the path of the recoater blade, causing collision \cite{recoater-monitor-2018}. Therefore, prediction of the z-displacement field can aid in determining whether a blade collision failure will occur. Because LPBF simulations can take minutes or even hours, using a surrogate model to predict the results instantly will significantly accelerate design for AM. For this reason, we train our models to predict the z-displacement at every node in the part at the final time step.

Of the 19,732 successful simulations, 16,594 shapes are used for training the model. The remaining 3,138 shapes are set aside during training and later used for testing. Each mesh consists of thousands of nodes and edges. This dataset covers a wide range of shapes, with variation between shapes ranging from subtle to drastic; it is therefore large enough to train a strong data-driven field prediction model. A figure containing a gallery of sample shapes and results from this dataset can be found in the Appendix. Note that the shapes and fields used in the 3-D dataset are distinctly different from those in the 2-D dataset; we aim to show that our model is capable of strong performance in multiple engineering contexts.

To maximize value to the scientific community, the dataset adheres to the FAIR guiding principles for datasets \cite{fair-2016}. First, it is \textit{findable}: metadata is informative and the dataset file structure is well-documented. Humans and computers alike can navigate the dataset. The dataset is also \textit{accessible}: The dataset is freely accessible at \url{https://github.com/kevinferg/graph-field-prediction}. Furthermore, metadata is included separately in the repository so that a full dataset download is not necessary to view a summary of the simulation results. Our dataset is \textit{interoperable} in that it is stored as compressed NumPy \texttt{.npz} files, and Python functions to read and visualize the data are included. Finally, it is \textit{reusable}: the dataset and accompanying code are released under the MIT license, and the dataset generation details are thoroughly documented.


\subsection{Topology-Agnostic Graph U-Net}

As previously discussed, to effectively transmit long-range information in a GNN, several layers of convolution must be stacked in series, which can quickly lead to vanishing gradient problems and is inefficient. We propose Topology-Agnostic Graph U-Net (TAG U-Net), which adds pooling and un-pooling steps in an encoder-decoder structure akin to a U-Net. The proposed TAG U-Net Architecture is shown in Fig.\ref{fig:architecture}.

\begin{widefigure}[t]
    \centering
    \includegraphics[width=\textwidth]{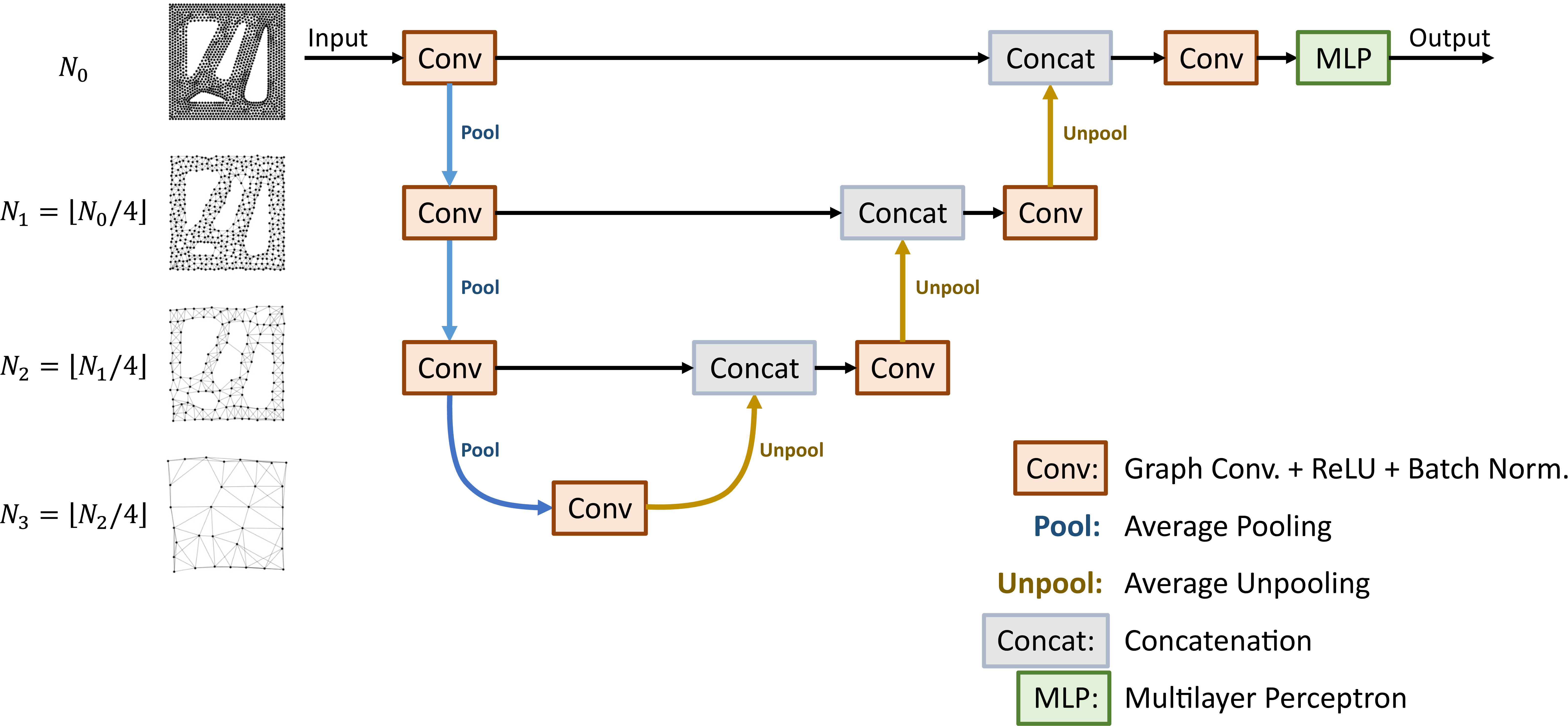} 
    \caption{Topology-Agnostic Graph U-Net (TAG U-Net) Architecture}
    \label{fig:architecture}
\end{widefigure}


As seen in Fig.\ref{fig:architecture}, the input mesh is first passed to the network as a collection of nodes whose features are their 2D x-y (or 3D x-y-z) coordinates and a set of bi-directional edges connecting neighboring nodes for use in convolution. Note that the input coordinates are sent through a linear transformation to expand the dimensionality of nodal features before being passed to the convolutions.

Next several convolution, activation, normalization, and pooling operations are performed on the transformed input as follows: An EdgeConv convolution operation is applied first, followed by ReLU activation and batch normalization. $k$-d tree pooling is then applied to the graph, reducing the graph's resolution before again performing a separate convolution, activation, normalization, and pooling. These steps repeat until the lowest graph resolution is reached. Note that the input to each pooling step is preserved for use in skip-connections.

Convolution, activation, and batch normalization at the lowest resolution are succeeded by $k$-d tree un-pooling, beginning the upward leg of the Graph U-Net. Before each of these convolutions, the stored nodal feature map (corresponding to the same mesh resolution from earlier in the network evaluation) is concatenated to the feature vector. This skip-connection preserves high-resolution geometric information across the U-Net bottleneck. The choice of EdgeConv for convolution and the proposal of $k$-d tree pooling are justified in the following sections.

\subsubsection{Convolution}

In standard 2D and 3D image/grid settings, convolution can be performed by locally multiplying a filter element-wise and summing the result at every pixel/voxel. At each stage of a convolutional neural network, the result of a convolution is a feature map, consisting of several feature channels at each pixel/voxel.

For spatial GNNs, convolutions use ``message-passing" to exchange feature information between neighboring nodes. Depending on the choice of convolution, a specific function is applied to combine these features, and they are aggregated into a single set of features. Such a process occurs at each node, and the result is an output feature map: a vector of features at each node, comparable to the vector of features at each pixel after and image convolution \cite{gcn-survey-2021}

Because the spatial location of each node in a mesh-based graph is similar in meaning to the location of a pixel/voxel in an array, the ideal analog of image convolution is therefore a general convolution that performs a similar vector-to-vector mapping at each node. Several such methods exist, including those nicknamed $\mathcal{X}$-Conv \cite{xconv-2018}, PointConv \cite{pointconv-2020}, and EdgeConv \cite{edgeconv-2019}. In its original implementation, EdgeConv is used as a general convolution for predictions on dynamic point clouds; we propose using EdgeConv as the convolution operation in our TAG U-Net. It is defined as

\begin{equation}
    \boldsymbol{x}_i' = \max_{j \in \mathcal{N}(i)} h_{\boldsymbol{\Theta}}(\boldsymbol{x}_i\; ||\; \boldsymbol{x}_j - \boldsymbol{x}_i),
\end{equation} 

where $\boldsymbol{x}_i'$ is the new (post-convolution) vector of features at node $i$. $\boldsymbol{x}_i$, the current (pre-convolution) feature vector at node $i$, is concatenated to $\boldsymbol{x}_j - \boldsymbol{x}_i$, the difference between features at node $i$ and those at node $j$, for all neighbors of node $i$, $j \in \mathcal{N}(i)$. These concatenated vectors are passed through an MLP $h_{\boldsymbol{\Theta}}$, in which $\boldsymbol{\Theta}$ are trainable parameters. 

The intent behind EdgeConv is to have a fully general function that considers a node's features relative to those of its neighbors and produces a set of features that can characterize the node within its local context. The MLP and maxpool operations allow significant flexibility in what kind of relationships each convolution is able to learn. Because we always expect node location to be a relevant feature, we concatenate node coordinates to the set of input features prior to each convolution. This is not mandatory, but we observed improvements by doing this in early tests. This step was also performed in the baseline models for fair comparison.


We also wish to show how a less complex convolution is able to perform on the problem, to establish the benefit of a convolution like EdgeConv in this context. One less elaborate graph convolution, GCNConv, was established by Kipf and Welling \cite{GCNConv-2017}, and can be expressed as
\begin{equation}
    \boldsymbol{x}_i' = \boldsymbol{\Theta}^\top \sum_{j \in \mathcal{N}(i) \cup \{i\}} \frac{\boldsymbol{x}_j}{\sqrt{\hat{d}_i\hat{d}_j}},
\end{equation}
where $\boldsymbol{\Theta}$ is a trainable feature matrix, which is transposed and multiplied by the element-wise sum of feature vectors of each node and its neighbors, normalized with respect to their degrees in the graph $\hat{d}_i$ and $\hat{d}_j$. We will also demonstrate the performance of our Graph U-Net using this convolution, treating it as a baseline. We posit that the implicit featurization of each connected pair of nodes via $\boldsymbol{x}_j - \boldsymbol{x}_i$ in EdgeConv makes it more suitable for use in a geometric context. Furthermore, its built-in MLP gives EdgeConv the expressive power of a neural network, rather than the much more restricted linear matrix multiplication. We therefore expect the TAG U-Nets with EdgeConv to have better performance than those with GCNConv as their convolution method.

\subsubsection{Pooling} 
\label{sec:pooling}
We propose a novel approach to coarsening graphs using a truncated $k$-d tree of nodes in n-dimensional space. This coarsening strategy corresponds naturally to simple pooling and unpooling operations, making it ideal for use in Graph U-Net models. 

Mesh coarsening, or ``decimation" strategies often seek to preserve mesh characteristics with graphical importance; that is, they reduce redundancy and prioritize preserving key visual features \cite{decimation-1998}. Some methods are better suited to geometric computation contexts, such as edge contraction, which attempts to minimize element aspect ratio \cite{edge-contraction-2003}.

\begin{figure}[H]
    \centering
    \includegraphics[width=0.48\textwidth]{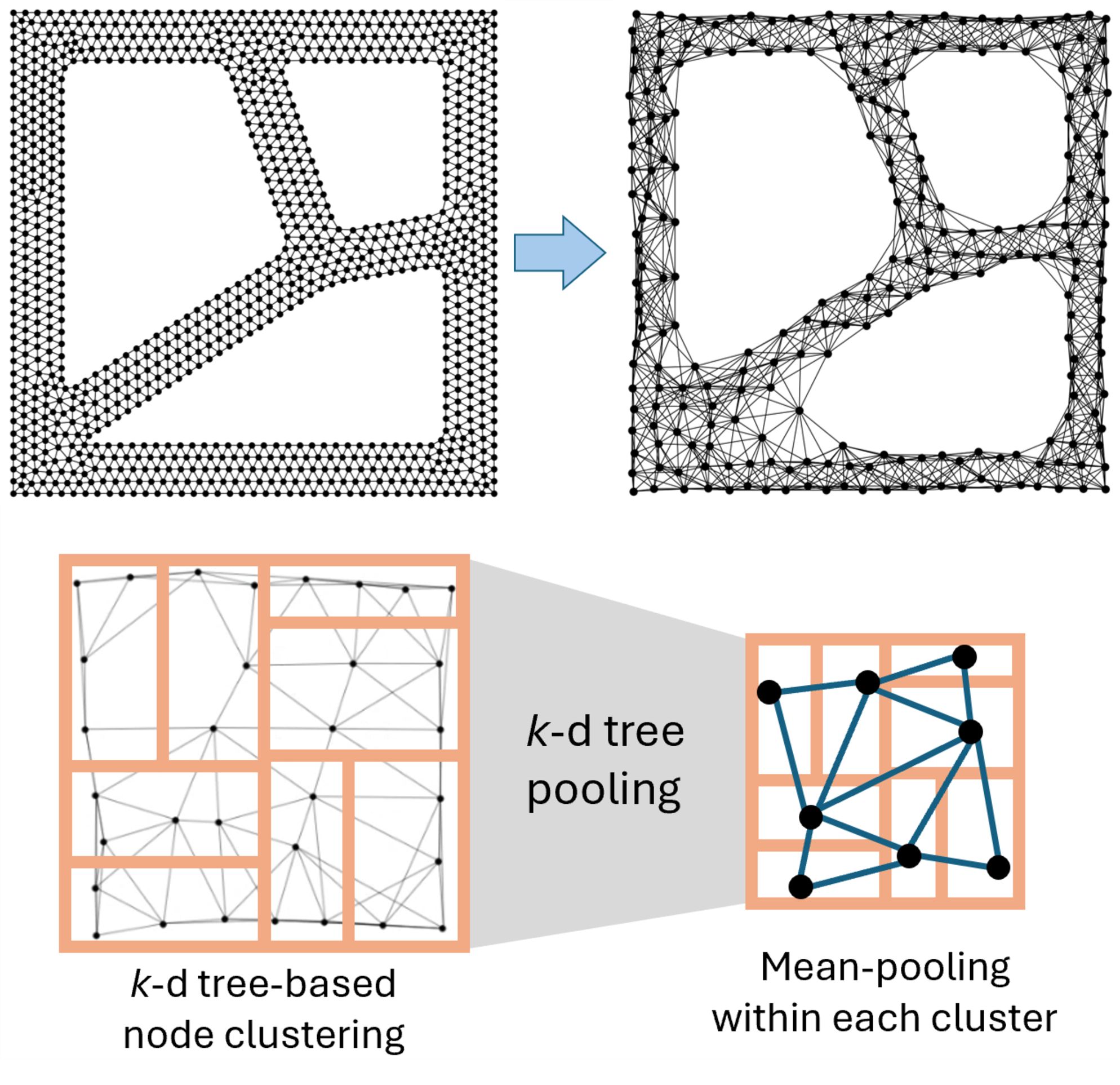}
    \caption{$k$-d tree pooling, demonstrated for a 2-D mesh}
    \label{fig:pooling}
\end{figure}

However, in Graph U-Nets, preserving these characteristics should be secondary to preserving other quantities that are of greater relevance in graph networks, such as local node density. We propose $k$-d tree pooling, depicted in Fig.\ref{fig:pooling} to accomplish this. To coarsen a graph using $k$-d tree pooling, we first generate a $k$-d tree for the collection of all nodes in the original graph. The tree generated is balanced, so it separates the given Euclidean space into two halves such that 50\% of the nodes are in each half-space. Performing this split recursively with each new half results in generating many cells, each containing 1 node. For our proposed method, we truncate the process, preventing further splitting once there are 4 nodes remaining in a cell (for the 2-D case). This results in clusters of 4 nodes; the average location of the nodes in a cluster is the location of a single node in the coarsened graph. The choice of 4 nodes per cell is in analogy with $2\times2$ image pooling with stride 2, in which groups of four pixels are pooled into a single pixel on a coarsened image. For the 3-D case, in keeping with the image pooling analogy, we use 8-node clusters, corresponding to the 8 pixels that are pooled in $2\times2\times2$ stride 2 pooling.

To pool a feature map from a graph to its coarsened version, the feature vectors of each node in a cluster are averaged element-wise, giving a mean-pooled feature vector for each cluster. This is accomplished using the PyTorch Geometric function \texttt{avg\_pool\_x()} \cite{pyg-2019}. To un-pool on the decoder side of the network, the feature vector at the coarsened node is copied to each of the constituent nodes within its cluster. Gradient information is preserved as expected for both the pooling and un-pooling operations.

To properly use the coarsened version of the graph in convolution in the deeper layers of the Graph U-Net, edges must be assigned in the coarsened graph. Establishing this edge connectivity can be a complicated task, but we found good results by simply constructing a $k$-nearest neighbors graph in the coarsened space. That is, for each node $i$, we assign a bi-directional edge between node $i$ and the $k$ nodes with the smallest Euclidean distance to node $i$. A value of $k=12$ worked well for our networks, but further optimization of this value, or another algorithm to re-assign edges, may be worth developing in the future.


\subsection{Training Information}

We select different architectures for the 2-D and 3-D models. In practice, the optimal hyperparameters of the model will depend upon the problem and the resolution of shapes in the dataset. Consider the model and training details listed here to be a starting recommendation for 2-D and 3-D problems on datasets with shapes of similar mesh resolutions. Note that the loss function is Mean Squared Error (MSE) across nodes in a batch of meshes, i.e.:

\begin{equation}
    L_{MSE}(\boldsymbol y, \boldsymbol f)  = \frac{1}{n}\sum_{i=1}^n \left( y_i - f_i \right)^2
\end{equation}

where $y_i$ is the $i$th element of ground truth field values $\boldsymbol y$, $f_i$ is the $i$th element of predicted field values $\boldsymbol f$, and $n$ is the total number of nodes in the batch of meshes. We benchmark each TAG U-Net model against two additional models: a plain GNN without the U-Net mechanism (e.g. omitting the pooling/unpooling steps and skip-connections, resulting in a chain of successive EdgeConv convolutions and an MLP), and a TAG U-Net with GCNConv convolutions instead of EdgeConv.

\subsubsection{Model details: 2-D problem}
For the model trained on the 2-D dataset, shapes were coarsened into clusters of 4 nodes, and the network contains 3 layers of pooling (and un-pooling). Each EdgeConv MLP contained 2 hidden layers of 128 neurons, used ReLU activation, and had 64 output channels. Convolutions were succeeded by ReLU and batch normalization. Concatenation was used for skip-connections. The output MLP had hidden layer sizes $[128, 128, 128]$ and ReLU activation, except at the last layer, which had no activation.

For training, the Adam optimizer with was trained with a learning rate of 0.001 for 75 epochs, which took 67 minutes on an NVIDIA GeForce RTX 2080 Ti GPU (11 GB DDR6 RAM). The batch size was 1 shape per batch.

\subsubsection{Model details: 3-D problem}
For the 3-D model, 8-node clusters were used for coarsening/pooling, and there were 3 pooling/un-pooling layers. Each EdgeConv MLP contained 2 hidden layers of 128 neurons, used ReLU activation, and had 128 output channels. As with the 2-D network, ReLU and batch normalization followed each convolution and concatenation skip-connections were used. The output MLP had 3 hidden layers with 256 neurons each and used ReLU activation, except at the last layer.

For training, the Adam optimizer with was trained for 50 epochs with a fixed learning rate of 0.001. The batch size was 4 parts, and training using one NVIDIA GeForce RTX 2080 Ti GPU (11 GB DDR6 RAM) took approximately 18 hours.

\section{Results and Discussion}
\label{sec:results}
In this section we present results of our analyses for the 2-D and 3-D problems. We tabulate and plot the $R^2$ results, comparing three GNN architectures: a plain GNN with EdgeConv, TAG U-Net with GCNConv, and TAG U-Net with EdgeConv. Next, we visualize the field prediction on typical shapes for both prediction tasks. We then investigate whether a trained TAG U-Net can be used in practice to classify nodes. Last, we perform a parametric study for the 3-D z-displacement prediction task, examining how model performance scales with the number of trainable parameters in the model.

\subsection{R-Squared Analysis}

To evaluate our model, we use the $R^2$ score, also referred to as the coefficient of determination \cite{r2-2005}. Each $R^2$ represents the goodness-of-fit for a single shape, comparing nodal predictions for each node in the shape with their ground-truth values. An $R^2$ of 1 denotes a perfect match between the prediction and the ground truth for all nodes in the shape. A value less than zero indicates a fit worse than if the mean response value were predicted for every input; when plotted as a scatter plot of predicted vs. ground truth nodal values, an ideal result would contain points lying close to the $y=x$ line, corresponding to an $R^2$ near 1. Table \ref{tab:r2-results} summarizes the median $R^2$ performance of each model on training and testing data for both the 2-D and 3-D problems.

{\renewcommand{\arraystretch}{1.1}%
\begin{widetable}[htbp]
    \centering
    \caption{Median $R^2$ values across training and testing shapes for 2-D stress prediction and 3-D z-displacement prediction tasks. Three GNN architectures are compared: a plain GNN with EdgeConv, TAG U-Net with GCNConv, and TAG U-Net with EdgeConv.}
    \begin{tabular}{V{3} c V{3} c | c V{3} c | c V{3}} \Xhline{3\arrayrulewidth}
    & \multicolumn{2}{c V{3} }{\textbf{2-D Stress Prediction}} & \multicolumn{2}{c V{3} }{\textbf{3-D Disp. Prediction}} \\
    \multirow{1}{*}{\textbf{Model}} & \multicolumn{2}{c V{3} }{Median $R^2$}& \multicolumn{2}{c V{3} }{Median $R^2$}\\ \cline{2-5}
        & Training & Testing & Training & Testing\\ \Xhline{3\arrayrulewidth}
        Plain GNN (w/ EdgeConv) & 0.439 & 0.415 & 0.664 & 0.649\\
        TAG U-Net (w/ GCNConv) & 0.544 & 0.540 & -0.102 & -0.065 \\
        TAG U-Net (w/ EdgeConv) & 0.901 & 0.874 & 0.874 & 0.855\\ \Xhline{3\arrayrulewidth}
    \end{tabular}
    \label{tab:r2-results}
\end{widetable}
}

For the 2-D stress prediction task, as seen in Tab.\ref{tab:r2-results}, we observe that the TAG U-Net with EdgeConv had the best performance, with a median $R^2$ of 0.87 on unseen testing data. The TAG U-Net that used GCNConv did not perform nearly as well on training or testing data, and the plain GNN with EdgeConv had the worst performance. None of the models exhibited significant overfitting, with the median testing $R^2$ values being only slightly lower than the median training $R^2$s in all cases.  

Table \ref{tab:r2-results} also tabulates training and testing results for the 3-D displacement prediction problem. Note that again the TAG U-Net with EdgeConv gives the best results, with a testing $R^2$ of 0.855. Once more, overfitting is minimal. For this problem, however, the plain GNN outperforms the GCNConv TAG U-Net, as the GCNConv TAG U-Net fails to give good results.

\begin{figure}[H]
    \centering
    \subfloat{{\includegraphics[width=0.48\textwidth]{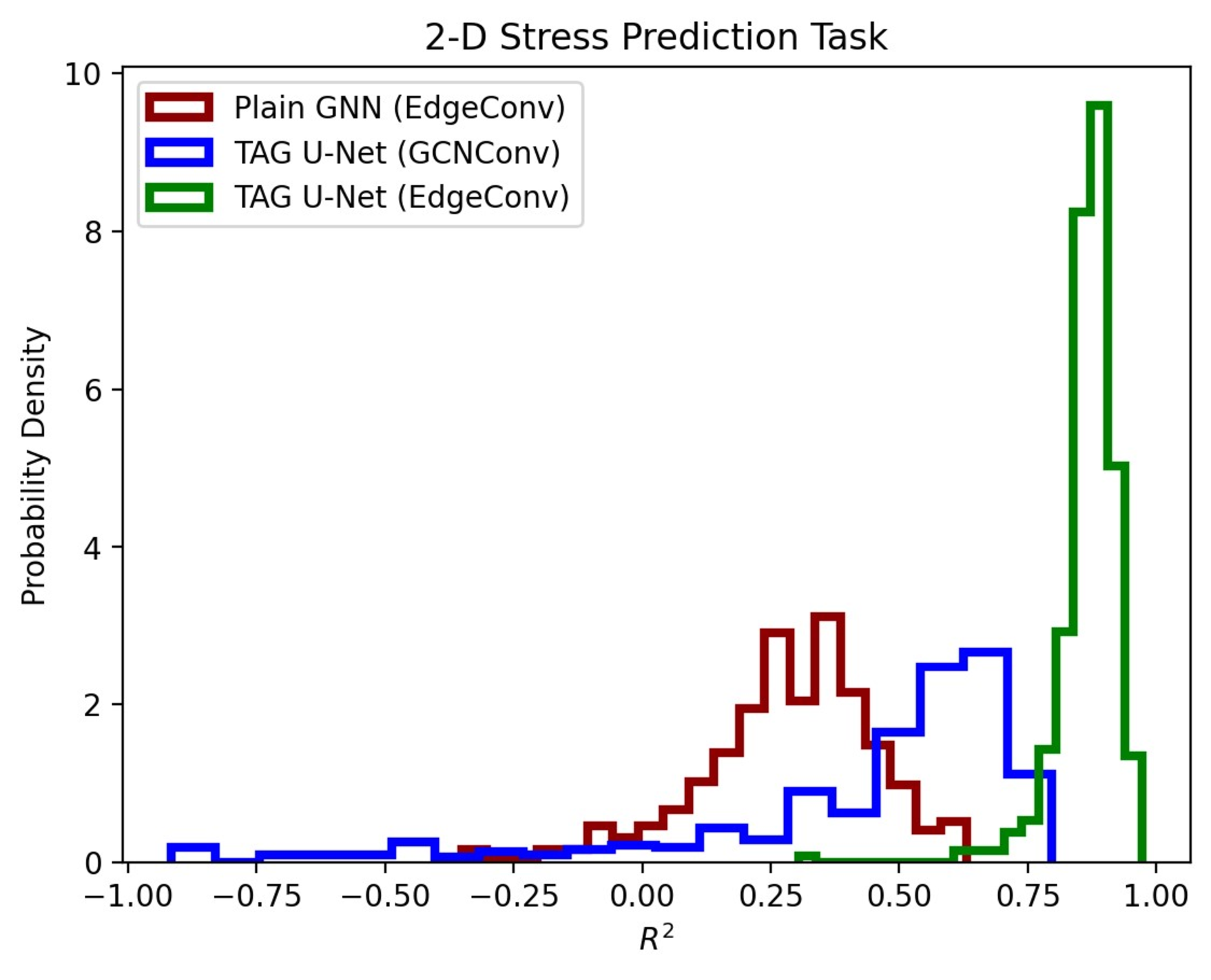} }}%
    \quad
    \subfloat{{\includegraphics[width=0.48\textwidth]{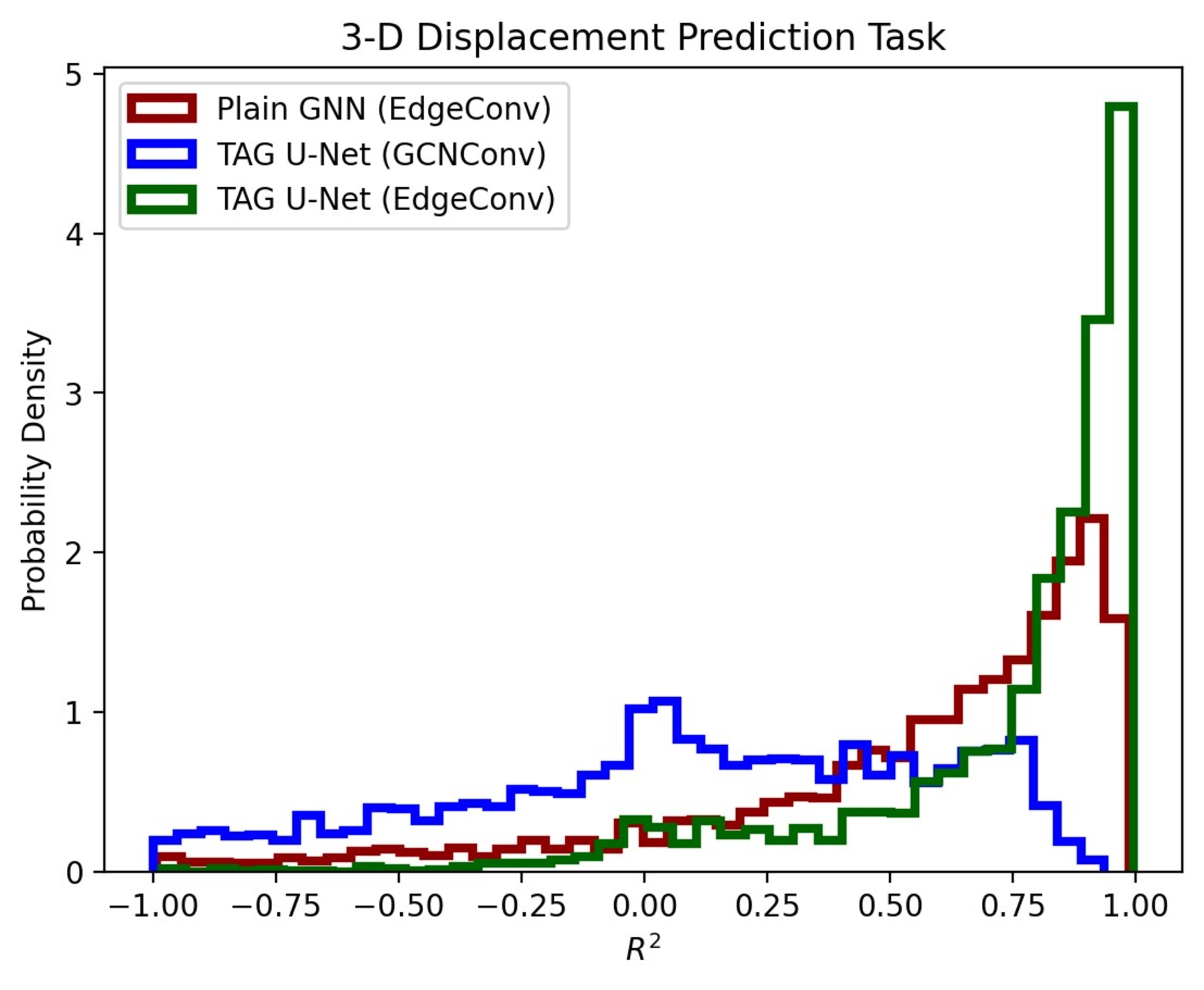} }}%
    \caption{$R^2$ distributions for test shapes in the 2-D stress prediction and 3-D displacement prediction tasks.}%
    \label{fig:histogram}
\end{figure}

The median values suffice to communicate how the models will perform on a ``typical" shape. However, the full distribution of $R^2$ values across the dataset will better elucidate the performance of each model. Figure \ref{fig:histogram} depicts the $R^2$ distributions as probability density histograms across \textit{testing} shapes for all three models on both prediction tasks.

The distributions in Fig.\ref{fig:histogram} reveal that TAG U-Net with EdgeConv out-performs its alternatives, with $R^2$ distributions centered at larger $R^2$ values, 
Note the left-skewness of all distributions seen in the figure; this is due in part to the notion that $R^2$, while capable of taking any negative value, cannot by definition exceed 1. 

We hypothesized that TAG U-Net's pooling, un-pooling, and skip-connections would make it superior to a plain GNN at field prediction tasks. Table \ref{tab:r2-results} and Fig.\ref{fig:histogram} provide evidence that corroborates this conjecture. We further hypothesized that GCNConv would not perform as well as EdgeConv, due to its simplicity in comparison, lacking both an edge featurization step and an MLP. Once again, investigating typical $R^2$ values and the distribution of $R^2$ throughout the dataset supports this hypothesis.

\subsection{Visualizations}

In this section we visualize TAG U-Net's predictions alongside the ground truth and benchmark models. We have selected the 75th-percentile test set shape for visualization on both the 2-D and 3-D problem. Figure \ref{fig:2d-comparison} displays 2-D model predictions, while 3-D predictions are shown in Fig.\ref{fig:3d-comparison}. The figures also contain plots of predicted vs. actual values, showing each node as a point on a scatter plot.

\begin{widefigure}[htbp]
    \centering
    \includegraphics[width=\textwidth]{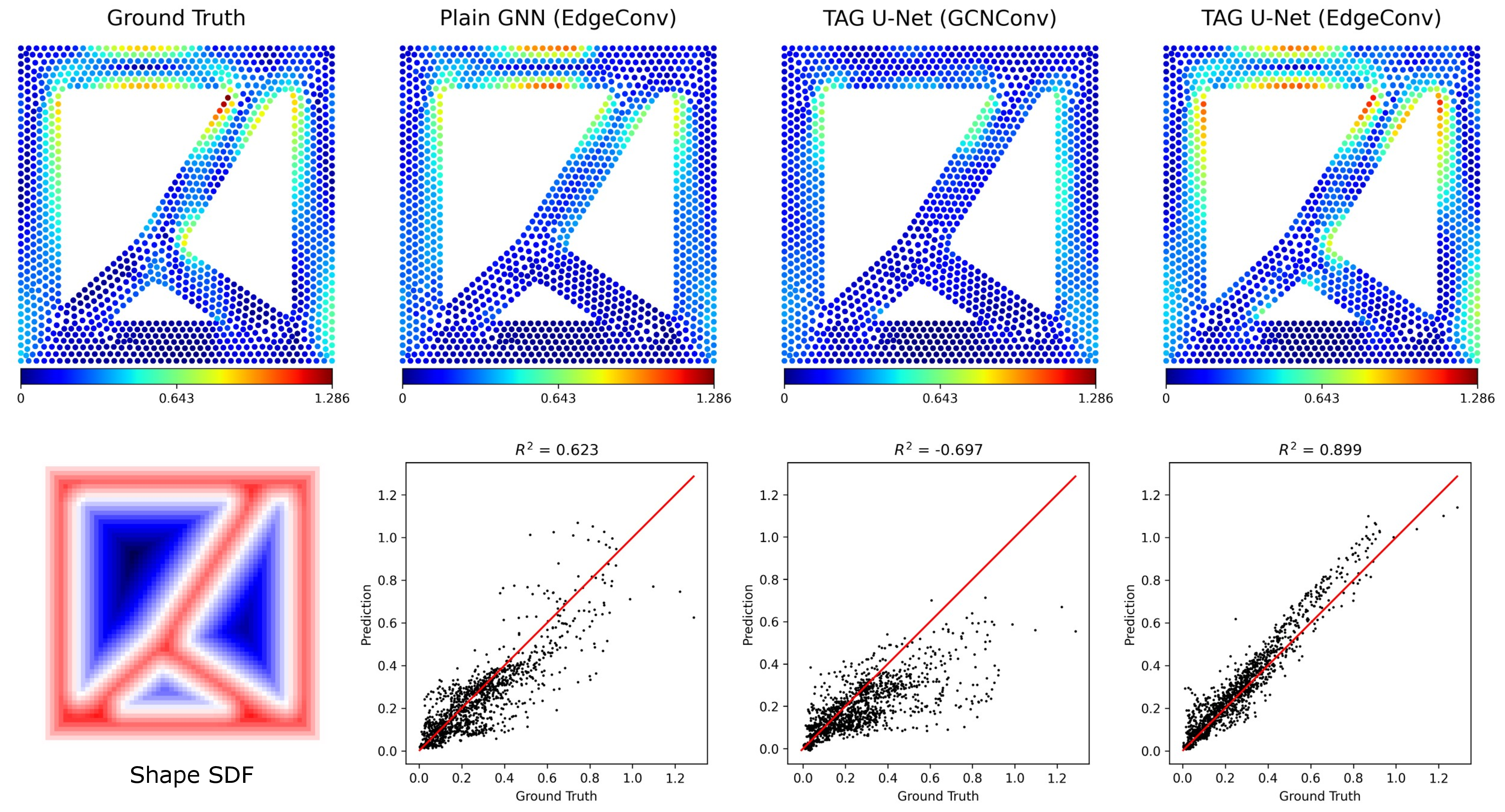}
    \caption{Visualizations of each model's predictions on a shape in the test set for the 2-D stress prediction problem. $R^2$ plots are also shown for each model. Units are $10^4 \times \text{Pa}$.}
    \label{fig:2d-comparison}
\end{widefigure}

As seen in Fig.\ref{fig:2d-comparison}, there is strong agreement between the ground truth and predicted fields. Peaks in the (EdgeConv) TAG U-Net stress field are in the same locations as, and have magnitudes very similar to, the ground truth. Note that while the Plain GNN can capture the areas of peak stress, in the shape shown it fails to capture the subtle variations in stress along the supports. In contrast, TAG U-Net, which is better suited to characterize long-range geometry, accurately captures the stress field, even for non-peak areas, giving it better value as a surrogate model for stress prediction.

The predicted vs. actual plots provided in the figure show a cluster of points tightly packed around the $y=x$ line, demonstrating that the nodal predictions tend to match closely with their ground truth values uniformly throughout the shape for the best model's prediction. Notice that the Plain GNN and GCNConv TAG U-Net have more scattered points in the $R^2$ plots, which is indicative of a less powerful predictive model. TAG U-Net with EdgeConv does appear to make slight over-predictions, particularly for high-stress regions. While this did not occur in every shape, whenever it does occur, it may be attributed to the relative rarity of high-stress nodes compared to nodes of low/medium-stress. In this case, the prediction is still very strong, but ensuring a training dataset contains enough of this ``rare" data in is essential for accurately predicting scalar fields, especially for unseen shapes.

\begin{widefigure}[htbp]
    \centering
    \includegraphics[width=\textwidth]{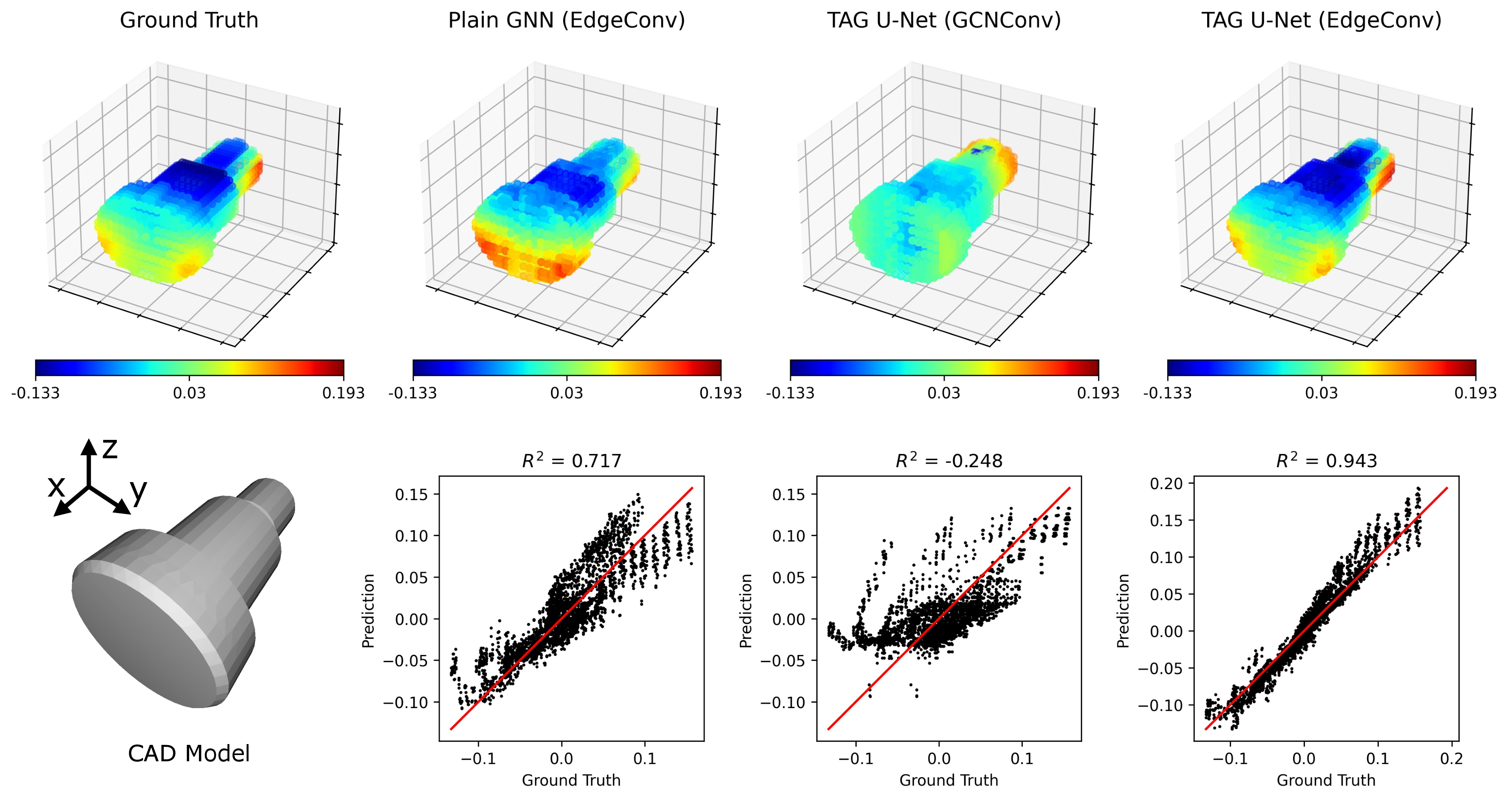}
    \caption{Visualizations of each model's predictions on a shape in the test set for the 3-D z-displacement prediction problem. $R^2$ plots are also shown for each model. Units are mm. Note that the z-direction is `up' on each 3-D plot.}
    \label{fig:3d-comparison}
\end{widefigure}

Figure \ref{fig:3d-comparison} demonstrates the same conclusion in 3-D; TAG U-Net with EdgeConv outperforms the other models. Note that although all three models roughly predict the correct location of peak displacement, the $R^2$ plots below show that EdgeConv TAG U-Net has the best fit with the ground truth. This agrees with the results from Tab.\ref{tab:r2-results} and further verifies our claims that the TAG U-Net architecture is especially effective at predicting arbitrary fields on unstructured geometric mesh data.

One additional observation we made that is not apparent in the figure is that when making predictions on the 3-D dataset, several shapes resulted in anomalously low $R^2$ values. Most often, these were shapes with extremely thin features containing high displacement values that the model had not learned to predict. As discussed, sufficient presence of anomalous data in the training set is critical for the deployment of surrogate models for design. Similarly, development of a system to detect such cases directly for a query shape may be worth investigating when implementing these models in real-world engineering applications.


\subsection{Additional Studies}

In sections, we narrow our focus on the 3-D AM z-displacement field prediction problem and investigate two further topics: a.) Whether scalar field prediction models can feasibly be interpreted as node-classification models, and b.) how much model performance improves with as network size changes. We feel both of these questions are essential to studying the value of a scalar field prediction model in a practical engineering setting.

\subsubsection{Case Study: Classification}
\label{sec:classification-study}
We trained our model to predict the z-displacement of nodes throughout an additively manufactured part. In practice, field predictor models are highly useful at determining where quantity of interest surpasses a threshold. For example, for a part manufactured via LPBF, if z-displacement is too high \cite{recoater-characterization-2022}. For the purpose of demonstration, we assign a threshold of 0.02 mm and take any z-displacement predictions above this to be potential areas of concern. Figure \ref{fig:disp-visualization} displays thresholded model predictions for each of the models discussed in the previous section, for two near-median shapes selected at random from the AM test dataset alongside the ground truth.

\begin{widefigure}[htbp]
    \centering
    \includegraphics[width=\textwidth]{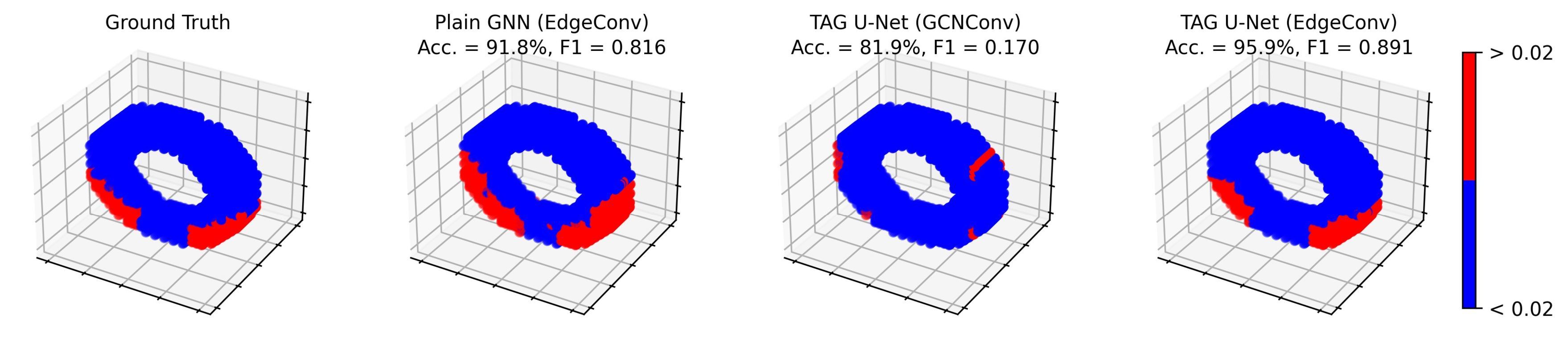}
    \includegraphics[width=\textwidth]{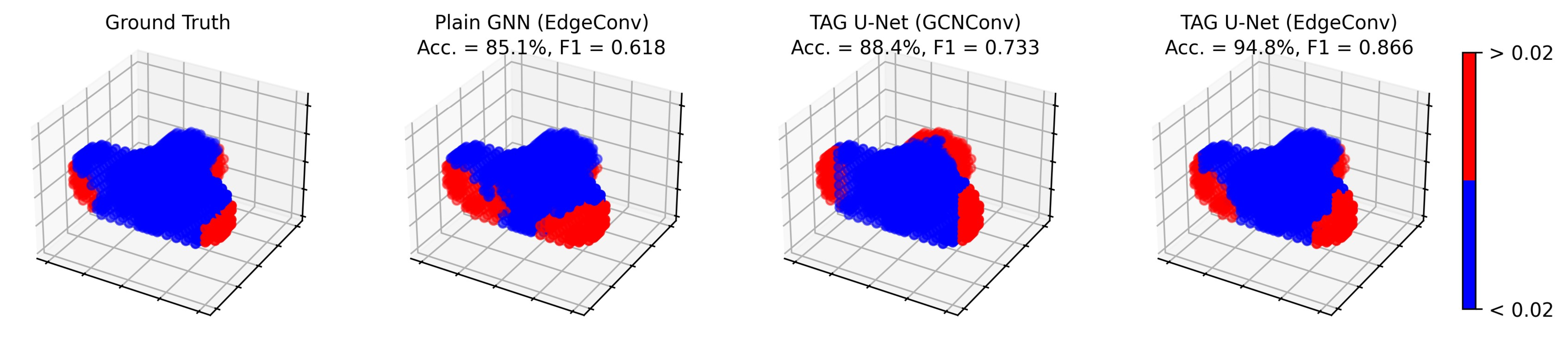}
    \caption{Visualization of each model's performance on two random shapes in the 3-D z-displacement AM test dataset, when used to classify nodal z-displacements as above/below a threshold of 0.02 mm. Accuracy and F1 score values are reported. Note that the z-direction is ‘up’ on each 3-D plot.}
    \label{fig:disp-visualization}
\end{widefigure} 

As seen in Fig.\ref{fig:disp-visualization}, there are both qualitative and quantitative similarities between the predicted and ground truth displacement threshold fields. Treating the regression models as classifiers in this way gives good accuracy, especially for the EdgeConv variant of TAG U-Net, which had a median test accuracy of 95.0\% and a median test F1 score of 0.842, meaning the model typically has strong performance as a threshold-based classifier. For comparison, the median test accuracy and F1 score were 92.2\% and 0.766 for the plain GNN model, vs. 81.9\% and 0.527 for the GCNConv TAG U-Net, indicative of overall weaker classification performance compared to TAG U-Net with EdgeConv. If an AM part designer were to use a TAG U-Net model in this way, they could obtain rapid feedback on which sections of the part may require adjustments without needing to run a build simulation. The instantaneous results can maximize the productivity of a design team, as long as a high-fidelity simulation is used to verify the results of a more finalized design. 









\subsubsection{Parametric Study: TAG U-Net Size} 
\label{sec:parametric-study}

The TAG U-Net framework has demonstrated strong performance on both 2-D and 3-D problems. However, one aspect of the model we wish to explore further is the extent to which the model size impacts its performance. Such a study is valuable for knowing whether increasing the capacity of the model will significantly increase its performance on test data or whether improvements will instead be marginal.

We therefore trained four TAG U-Nets across a range of sizes (trainable parameter counts) to determine both the parameter-efficiency of such networks and the extent to which overfitting becomes an issue as model size increases. For this parametric study, we focus on the 3-D z-displacement prediction problem. 

\begin{widetable}[htbp]
    \centering
    \caption{Performance of 4 models of Graph U-Net depth 3 on the 3-D z-displacement prediction task. Total parameter count, EdgeConv hidden layer sizes and output channels, and output MLP hidden layer sizes are listed, along with the resulting median $R^2$ on training and testing data.}
    \begin{tabular}{V{3} c V{2} c | c V{3} c | c | c V{3}}\Xhline{3\arrayrulewidth}
    \multicolumn{3}{V{3} c V{3} }{ Model Information } & \multicolumn{2}{c V{3} }{Median $R^2$}\\\Xhline{2\arrayrulewidth}
    Parameters & EdgeConv MLP & Output MLP & Training & Testing\\\Xhline{2\arrayrulewidth}
        28,385 & $(32, 32) \rightarrow 16$ & $(64, 64, 64)$ & 0.547 & 0.537\\
        100,737 & $(32, 32) \rightarrow 32$ & $(128, 256, 128)$ & 0.716 & 0.704\\
        341,377 & $(64, 64) \rightarrow 128$ & $(256, 256, 128)$ & 0.784 & 0.771\\
        630,785 & $(128, 128) \rightarrow 128$ & $(256, 256, 256)$ & 0.874 & 0.855\\
        \Xhline{3\arrayrulewidth}
    \end{tabular}
    \label{tab:parameters}
\end{widetable}

Table \ref{tab:parameters} contains the training and testing results for all four models, in addition to listing the total trainable parameter count and individual model details of each. Note that the largest model trained had the best training and testing performance, still without suffering significant overfitting. However, the performance improvement decreased with each subsequent model capacity increase. Therefore, choosing a network size is a decision that should be made by evaluating the tradeoff of model size and performance, also considering factors such as overfitting and training time. The results of simple network capacity studies like this should be interpreted with due skepticism, since it is possible that additional regularization can independently combat overfitting. However, such a study serves well as a general test of how the representation capabilities of a model scale with increasing network size.


\section{Conclusion and Future Work}
\label{sec:conclusion}

We demonstrated a TAG U-Net, a Topology-Agnostic Graph U-Net architecture that takes in an arbitrary graph representation of a 2-D or 3-D part, and makes a node-wise field prediction on that part. By training on simulation or finite-element analysis results for a sufficiently varied dataset of shapes, the proposed model can make good predictions on unseen shapes, with median $R^2$ values on testing datasets of approximately 0.87 and 0.82 for 2-D and 3-D shape datasets, respectively. The model makes use of EdgeConv as a convolution method, which was shown to perform significantly better than the simpler GCNConv. A proposed pooling approach, $k$-d tree pooling, provides a method for coarsening general graphs in a way that preserves local node density. When used in the proposed architecture, this pooling approach is effective; our resulting TAG U-Nets performed significantly better than comparable pure Graph Neural Networks (which lacked pooling/unpooling) for the same scalar field prediction tasks in both 2-D and 3-D.

Our method demonstrates the efficacy that graph learning methods can have in a geometric context. This means that shapes need not be transferred onto a grid domain, as traditional convolutional neural networks are not required. Instead, an arbitrary mesh structure is sufficient as input, and any target field in the training data can be predicted, provided the training dataset contains sufficient size and diversity.

We generated a full 3-D additive manufacturing dataset to test our method. The dataset contains Netfabb Laser Powder Bed Fusion nodal displacement results for over 19,000 parts from the Fusion 360 Gallery Segmentation Dataset. The dataset satisfies the FAIR principles for scientific datasets, and it has been made publicly available and can be easily manipulated in Python. Datasets like this are essential to the development of deep learning methods in engineering. Our dataset in particular can be used design for additive manufacturing; the displacement fields it contains indicate regions on the part where a build may fail due to recoater blade interference. The models trained in this work therefore demonstrate immediate applicability in this space, possessing the ability to identify which nodes in a given shape may fail with high accuracy. After the initial training process, TAG U-Net can make nodal predictions for an entire part in less than one second. If used in an iterative process, the time saved over running a full simulation at each iteration would be significant. We hope to demonstrate such a process in future work.

While EdgeConv proved highly effective, its reliance on large multilayer perceptron evaluations can be costly, especially when the node or edge count is high. Investigation into alternatives that streamline the computations required for each convolution should be considered. Similarly, while the proposed $k$-d tree pooling strategy works well, its use of $k$-nearest neighbors pooling to assign edge connectivity on the coarsened graph, while simple, may not provide the same benefits as more sophisticated methods that attempt to preserve connections from the previous resolution. The TAG U-Net architecture is highly modular, allowing convolution and pooling components to be easily substituted with alternatives; a comprehensive look at different combinations of these components would be highly valuable.

Last, models capable of predicting multiple fields, especially for a range of input parameters would be a natural extension of our model. Prediction of vector fields could be accomplished by simply increasing the size of the output dimension of the model. For example, the TAG U-Nets trained in this work can perform inference for only a single input problem: for the 2-D task, von Mises stress for a single set of 2-D loads and boundary conditions, or in 3-D, z-displacement for additive manufacturing of a build with one material using one set of process parameters. While these models operate on any arbitrary shape, a method to additionally encode additional inputs, such as variable material properties, loads, or process parameters, would significantly increase the utility of a given model.

\section*{Acknowledgments}
This research was funded by Air Force Research Laboratory contract FA8650-21-F-5803.


\newpage
\appendix

\section*{Appendix: Dataset examples}
\label{sec:appendix}

\begin{widefigure}[htbp]
    \centering
    \includegraphics[width=0.95\textwidth]{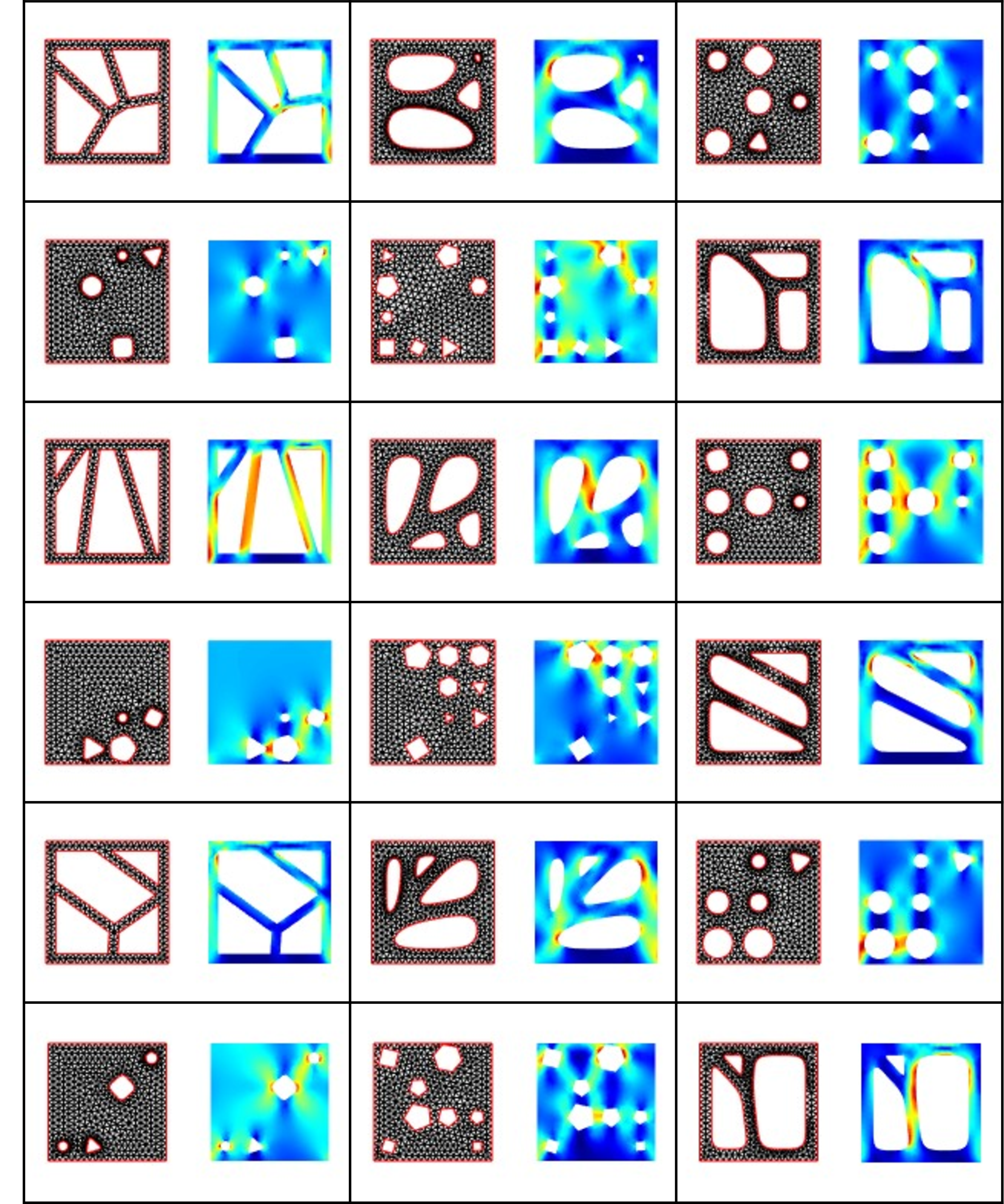}
    \caption{A random selection of entries in the 2-D dataset. Meshed shapes are displayed alongside output nodal von Mises stress fields}
    \label{fig:2d-samples}
\end{widefigure}

\begin{widefigure}[htbp]
    \centering
    \includegraphics[width=\textwidth]{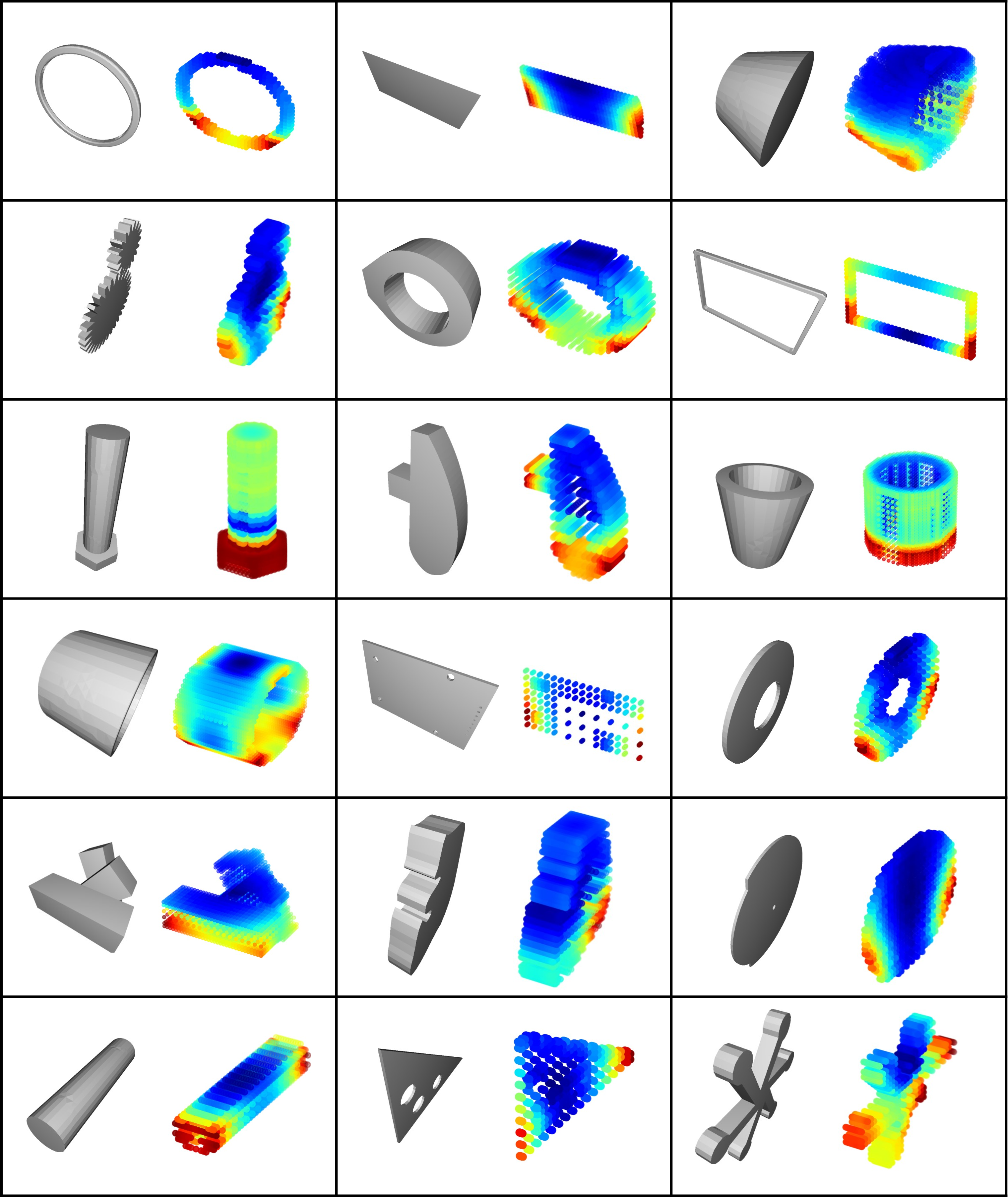}
    \caption{A random selection of entries in the 3-D dataset. Input CAD models are displayed alongside output nodal z-displacements plotted as point clouds}
    \label{fig:3d-samples}
\end{widefigure}

\end{document}